\pdfoutput=1

\documentclass[11pt]{article}

\usepackage[final]{acl}
\usepackage{color} 
\usepackage{bbm}

\newcommand{\C}[1]{\textcolor{black}{#1}}
\usepackage{times}
\usepackage{latexsym}
\usepackage{graphicx}  
\usepackage{adjustbox}
\usepackage[T1]{fontenc}
\usepackage{textcomp}

\usepackage[utf8]{inputenc}
\usepackage{sectsty}
\usepackage{graphicx}
\usepackage{multicol}
\usepackage{microtype}
\usepackage{enumitem}
\usepackage[linesnumbered,ruled,lined,noend]{algorithm2e}
\usepackage{listings}
\usepackage[frozencache=true,cachedir=minted-cache]{minted} 

\usepackage{microtype}

\title{Lost In Translation: Generating Adversarial Examples Robust to \C{Round-Trip Translation}}

\author{Neel Bhandari \\
  RV College of Engineering \\
  \texttt{neelbhandari64@gmail.com} \\\And
  Pin-Yu Chen \\
  IBM Research \\
  \texttt{pin-yu.chen@ibm.com} \\}

\begin{document}
\maketitle
\begin{abstract}
Language Models today provide a high accuracy across a large number of downstream tasks. However, they remain susceptible to adversarial attacks, particularly against those where the adversarial examples maintain considerable similarity to the original text. 
Given the multilingual nature of text, the effectiveness of adversarial examples across translations and how machine translations can improve the robustness of adversarial examples remain largely unexplored. In this paper, we present a comprehensive study on the robustness of current text adversarial attacks to round-trip translation. We demonstrate that 6 state-of-the-art text-based adversarial attacks do not maintain their efficacy after round-trip translation.
Furthermore, we introduce an intervention-based solution \footnote{Code for the paper: \url{https://github.com/neelbhandari6/NMT_Text_Attack}.\newline
Emails: Neel Bhandari: neelbhandari64@gmail.com \newline
Pin-Yu Chen: pin-yu.chen@ibm.com} to this problem, by integrating Machine Translation into the process of adversarial example generation and demonstrating increased robustness to round-trip translation. Our results indicate that finding adversarial examples robust to translation can help identify the insufficiency of language models that is common across languages, and motivate further research into multilingual adversarial attacks.
\end{abstract}

\section{Introduction}

Language models, despite their remarkable success across tasks, have shown to be vulnerable to adversarial examples, which are inputs designed to be similar to the model's native data inputs, but crafted with small modifications to fool the model during inference. These examples can be classified correctly by a human observer, but often mislead a target model, providing an insight into their robustness to adversarial inputs \cite{chen2023holistic,chen2023book}. \C{They are essential in understanding key vulnerabilities in models across a variety of applications \cite{chen2023ai}.}

\C{ML models are being increasingly deployed commercially for translation. A special form of translation is round trip translation, which focuses on translating a given text from one language to the second and back to the first. Round trip translation has been increasingly used in several research areas, including correcting grammatical errors \cite{grammar,grammar1}, evaluating machine translation models \cite{eval,eval1,eval2}, paraphrasing \cite{paraphrasing} and rewriting questions \cite{questions}. It is also used extensively as part of the quality assurance process in critical domains such as medical, legal and market search domains. The use of ML models in these critical domains means that they have to be tested by robust adversarial attacks to make for safe and reliable commercial deployment.
Given the importance of round trip translation, we are motivated to study its effects on current adversarial attacks. }

We summarise our contributions as follows:
\begin{itemize}

  \item \C{We demonstrate that round trip translation can be used as a cheap and effective defence against \textit{current} textual adversarial attacks. We show that 6 state-of-the-art adversarial text attacks suffer an average performance loss of 66\%, rendering most examples generated non-adversarial.}

  \item \C{However, we find that round-trip translation \C{defensive capabilities} can be bypassed by our proposed  \textit{attack-agnostic} algorithm that provides machine translation intervention to increase robustness against \C{round-trip translation}.}We find it provides minimal difference in quantification metrics to the original, which shows our method finds a new set of robust and high-quality text adversarial examples against neural machine translation (NMT). 
\end{itemize}

\section{Related Works}
\cite{papernot2017practical} proposed a white box adversarial attack that repeatedly modified the input text till the generated text fooled the classifier. This method, although effective in principle, did not maintain semantic meaning of the sentence.
\cite{ebrahimi-etal-2018-hotflip} and \cite{samanta2017crafting} proposed gradient-based solutions involving token based changes and searching for important words. These methods, however, did not prove to be scalable and lacked robust performance.
It was followed by methods such as character replacement \cite{ribeiro-etal-2018-semantically}, phrase replacement and word scrambling. These techniques, however, fail to maintain semantic consistency with the original input.
\cite{jia2019certified} introduced adding distracting sentences to the reading comprehension task. \cite{jin2020bert} propose TextFooler which generates adversaries using token-level similarity and is bound by axiomatic constraints. \cite{lei2018discrete} propose paraphrasing attacks using discrete optimization.
\cite{garg-ramakrishnan-2020-bae} introduce BAE, which uses masked-language modelling to generate natural adversarial examples for the text.
Recent works in adversarial attacks on NMT include \cite{cheng2019robust} using gradient based adversarial inputs to improve robustness of NMT models,
and \cite{zhang-etal-2021-crafting} proposed a novel black-box attack algorithm for NMT systems. However, none of these works target \C{round-trip translation}, and do not demonstrate attack agnostic capabilities.

\begin{algorithm}[t]
\caption{NMT-Text-Attack}
\label{algo_NMT}
\SetKwInOut{Input}{Input}
\SetKwInOut{Output}{Output}
\Input{Sentence $S=[w_1,w_2,..,w_n]$, Ground truth label $Y$, Victim Model $V$, Machine Translation model $M$, User-Specific Constraints $C$, Attack $A$}
\Output{Adversarial Example $X_a{}_d{}_v$}

\BlankLine
\textbf{Phase I - Word Importance Ranking} \\
Call attack A \\
Initialize edge weights \\
\For{each word $w_i$ in $S$}
{   
    Compute Importance score $I_i$ from $A$
}
Sort words in descending order into list $W$ \\
\textbf{Phase 2 - Word Replacement} \\
\# Word Replacement Strategy \\
\For{each word $w_i$ in $W$}
{
    Predict Top-K replacements for $w_i$ using $A$ and store in $R=[r_1,r_2,..,r_k]$ \\
    \For{each word $w_i$ in $W$}
         {
            Replace $w_i$ with $r_j$ in $S$ to make $X_a{}_d{}_v$ \\
            Round-Trip-Translate $X_a{}_d{}_v$ with $k$ language(s) using $M$ to make $T=[t_1,t_2,..,t_p]$ where $t_i$ is $X_a{}_d{}_v$ translated through language $i$\\
            Evaluate classification scores for $T=[t_1,t_2,..,t_p]$ using $V$, removing examples that do not maintain adversarial sentiment\\
            \For{each $c_i \in C$}
            {
            Apply constraint $c_i$ to each $t_i \in T$ \\
            }
            Select best $t_i \in T$ w.r.t constraints $C$ and store as $X_a{}_d{}_v$
         }
}
    \Return $X_a{}_d{}_v$
\end{algorithm}

\section{NMT-Text-Attack} 
In order to generate adversarial examples robust to round-trip translation, we propose an intervention-based attack-agnostic method that only requires access to a neural machine translation\C{(NMT)} model, \C{shown in Algorithm \ref{algo_NMT}}. We employ a generic template used by standard state-of-the-art adversarial attack examples in order to showcase the attack-agnostic capabilities. From \cite{Li2019TextBuggerGA,jin2020bert,ren-etal-2019-generating,garg-ramakrishnan-2020-bae,gao2018blackbox} it can be seen that the attacks follow a two section split. The first section is word importance ranking, and the second section deals with word replacement and constraint evaluation, where NMT-Text-Attack is introduced along with the original algorithm's constraints.

\textbf{I. Word Importance Selection.}\label{words}
This section initially involves pre-processing the input sentence with techniques such as removing stop words etc. This is followed by analysing the most important keywords in the target sentence using several techniques, ranging from the input deletion method, to probability weighted word saliency. These methods are specific to the adversarial attack chosen to be integrated with NMT-Text-Attack. For example, TextFooler uses the input deletion method. Once the most important words are learnt, attack algorithms look for replacements through synonym search or by 
\C{replacing individual characters of the original input word to make an adversarial candidate}. 
\SetKwInput{KwInput}{Input}              
\SetKwInput{KwOutput}{Output}

\textbf{II. Constraint Evaluation.}
\C{We introduce the machine translation task in this section. First, we predict the Top-K replacements for each word $w_i$ in word importance ranking list $W$ and substitute them in the sentence $S$ iteratively (Step 12). We then implement round-trip translation on these sentences for $k$ languages, where $k$ is specified by the user (Step 13). On collecting the candidate sentences, we evaluate them on the sentiment classification model $V$ and remove all examples that do not maintain the adversarial sentiment post round-trip translation (Step 14). Finally, we apply the algorithm-specific constraints $C$ on the collected final sentences $T$, and select the best candidate based on their similarity score with respect to the original sentence.} 

This is followed by applying algorithm-specific constraints $C$ such as semantic similarity to original input on replacement, POS tag preservation etc.

\section{Evaluation}
For performance evaluation, we consider using a range of algorithms from the TextAttack library \cite{morris2020textattack}.

\subsection{Dataset and Victim Model}
We use the Rotten Tomatoes Movie Reviews and Yelp Polarity datasets to perform sentiment analysis. We sample 1000 random examples from the test set of each of these mentioned datasets and run our experiments on them. For our Victim Model, we use the Bidirectional Encoder Representations from Transformers (BERT) model \cite{devlin2019bert}.

\begin{figure}[t]
\centering
\includegraphics[width=0.99\linewidth]{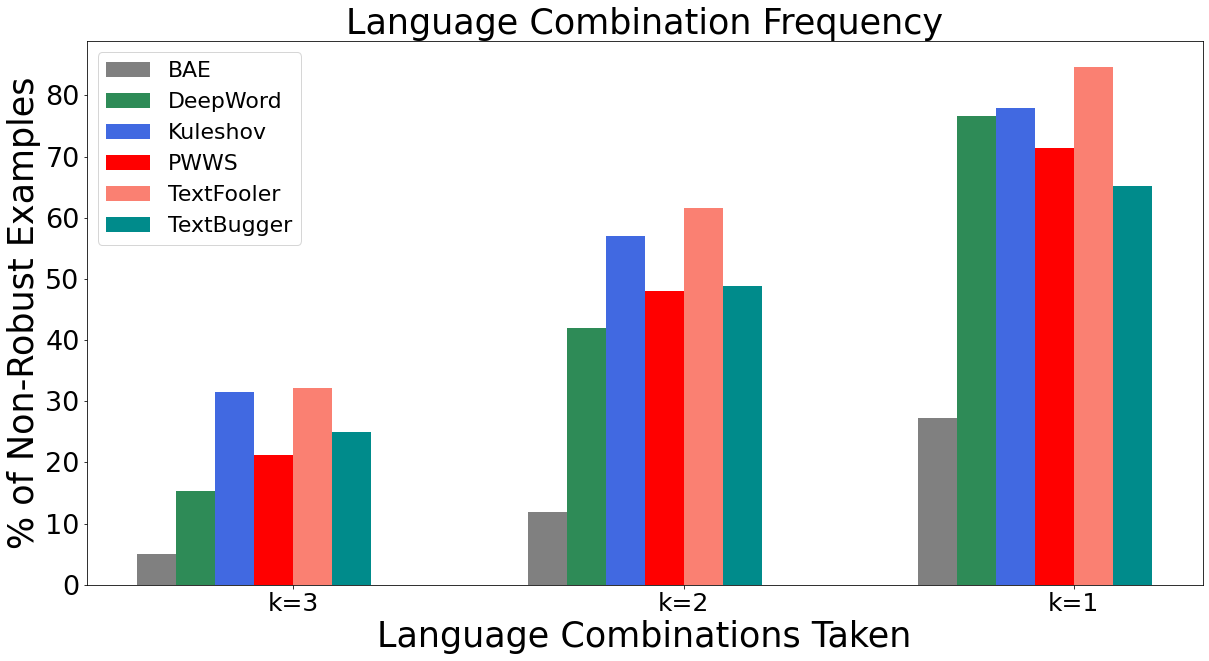}
\caption{Percentage of non-robust examples flagged by at least $k$ language combination}
\label{fig:rep}
\vspace{-2mm}
\end{figure}

\subsection{Current Attacks are not Robust to   \C{Round Trip Translation}} \label{sec3}
We run 6 adversarial attacks on the Movie Reviews Dataset and analyse their robustness to \C{round-trip translation}, as shown in Figure~\ref{fig:rep}. We analyse them against 3 languages -- Spanish, German and French through the EasyNMT library (see Appendix for more details). On  \C{round-trip translating} the adversarial examples, we test the resultant examples against the classification model. 

\C{On the y-axis, we provide the percentage of non-robust examples to at least $k$ out of $m=3$ languages. Formally, if $k$ is the number of languages used in tandem, $N$ is the number of examples in total, $y_a$ is the original prediction before round trip translation and $\hat{y_a}$ is the prediction after round-trip translation by translation model $M$ and victim model $V$, then the y-axis is defined as $Y=\frac{1}{N}\sum_{a=1}^N \mathbbm{1}{{\textrm{\{at least k languages have } y_a \neq \hat{y_a}\}}} $, where $\mathbbm{1}\{E\}$ is an indicator function such that it is one when the event $E$ is true and zero otherwise. }

We see that on average, over 66\% of the examples generated originally by the attack are rendered non-adversarial on  \C{round-trip translation} with at least one language ($k=1$). BAE remains the most robust to translations, while TextFooler remains the least robust. On increasing the number of language combinations taken ($k>1$), we see that there is a decrease in \C{effectiveness of round trip translation as a defense against the adversarial examples}, however there is still significant loss in attack success rate. \C{This is because when you add more languages as a constraint, there is an increased chance that at least one of the constrained languages is robust to round-trip translation for any example.}
\C{This provides considerable evidence that round trip translation can be used as a cheap and effective defense, and motivates the question of whether there exists text adversarial examples robust to round-trip translation}. In the following sections, We evaluate the robustness of our proposed NMT-Text-Attack as shown in Algorithm \ref{algo_NMT}.

\subsection{NMT-Text-Attack Results}

We analyse the results of incorporating NMT-Text-Attack into existing attacks across the mentioned datasets. We evaluate the attack on its success rate with respect to the attacks' native success rate without NMT-Text-Attack. Note that, through our novel intervention-based algorithm, we are able to guarantee 100\% robustness to back-translation on the user's selected language(s). This is because our algorithm (line 14) introduces a strict constraint to only allow examples that are robust to back-translation to be selected as candidates for the attack, which leads to significant increase over the original algorithm's robustness to  \C{round-trip translation}. This guarantee is important as it helps achieve high-quality robustness in multilingual settings, which no existing adversarial attack can provide. 
\C{Table~\ref{table:attack} shows that to meet this criteria, NMT-Text-Attack is successful on average 30\% less examples than it's original counterpart.}

\begin{table}[t]
\centering
\caption{Success Rate (\%) of NMT-Text-Attack Relative to when Original Attack Success Rate is 100\% (Replacement generation limit=40)}
\begin{adjustbox}{max width=0.99\columnwidth}
\begin{tabular}{l|lll} 
\hline
Dataset              & TextFooler+NMT & TextBugger+NMT & PWWS+NMT  \\ 
\hline
MR                   & 70.7       & 74.7       & 69.4    \\ 
\hline
Yelp                 & 60.0       & 71.4       & 68.8  \\ 
\hline
\multicolumn{1}{l}{} &            &            &      
\end{tabular}
\end{adjustbox}
\label{table:attack}
\vspace{-6mm}
\end{table}

\begin{figure}[t]
\centering
\includegraphics[width=0.99\linewidth]{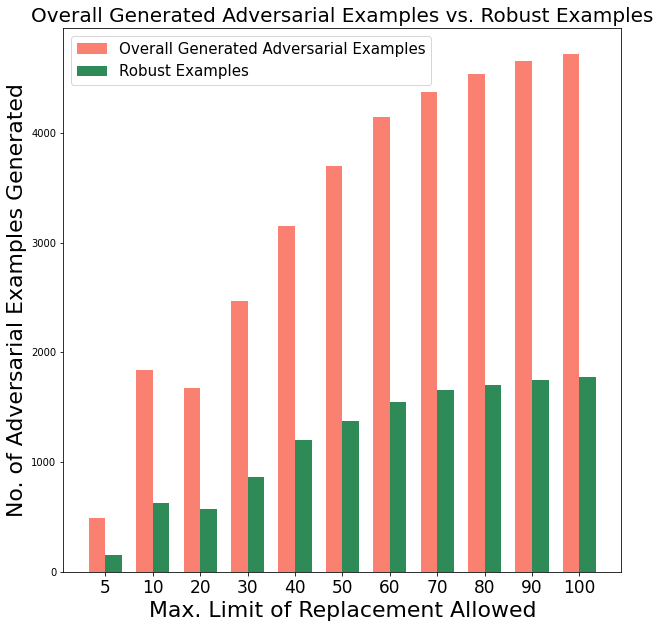}
\caption{Replacement vs. Robust Examples}
\label{fig:rep1}
\vspace{-2mm}
\end{figure}

While this \C{loss} may seem significant, we believe this is justified for two reasons. First, this loss comes with a 100\% success in robustness to  \C{round-trip translation} coupled with attack success. This is critical in commercial settings where deployed models need to have confident outputs in the face of several language translations. Secondly, in Figure~\ref{fig:rep1}, we see that there is considerable scope to increase the number of robust examples available simply by increasing the replacement limit. We set our replacement limit at 40 for our experiments, and Figure~\ref{fig:rep1} demonstrates that scaling the number of replacements significantly increases number of available robust examples.

\begin{table}[t]
\centering
\caption{Sentence similarity analysis on Yelp and Movie Reviews (MR) Datasets}
\begin{adjustbox}{max width=0.99\columnwidth}
\begin{tabular}{l|l|lll} 
\hline
Dataset      & Attack          & USE   & Jaccard & BERT    \\ 
\hline
Yelp          & TextBugger       & 0.93  & 0.79    & 0.95    \\ 
\cline{2-5}
              & TextFooler       & 0.93  & 0.81    & 0.97    \\ 
\cline{2-5}
              & PWWS             & 0.93  & 0.85    & 0.97    \\ 
\cline{2-5}
              & TextBugger + NMT & 0.94  & 0.848    & 0.9715    \\ 
\cline{2-5}
              & TextFooler + NMT & 0.82  & 0.724   & 0.956   \\ 
\cline{2-5}
              & PWWS + NMT       & 0.83  & 0.645   & 0.9265  \\ 
\cline{2-5}
\hline
MR            & TextBugger       & 0.93  & 0.79    & 0.95    \\ 
\cline{2-5}
              & TextFooler       & 0.813 & 0.715   & 0.953   \\ 
\cline{2-5}
              & PWWS             & 0.85  & 0.77    & 0.96    \\
\cline{2-5}
              & TextBugger + NMT & 0.91  & 0.68    & 0.92    \\ 
\cline{2-5}
              & TextFooler + NMT & 0.82  & 0.724   & 0.956   \\ 
\cline{2-5}
              & PWWS + NMT       & 0.83  & 0.645   & 0.9256  \\ 
\cline{2-5}
\hline
\end{tabular}
\end{adjustbox}
\label{table:perf}
\vspace{-2mm}
\end{table}

We also provide a quantitative analysis of our model by analysing the adversarial examples generated against the original attack in Table ~\ref{table:perf}. Universal Sentence Encoder ~\cite{cer2018universal} with cosine similarity, along with Jaccard Similarity are used as similarity metrics, while BERT Score ~\cite{bertscore} is used to analyse meaning preservation. We notice that there is little variation in the effectiveness of the algorithms when it comes to meaning preservation and similarity, which shows that our proposed intervention, while increasing robustness significantly, maintains the quality of the original attack. Examples of adversarial examples on sentences have been mentioned in the Appendix.

\subsection{Ablation Study} \label{ablation}

In this section, we provide an ablation study to substantiate the performance of our algorithm. In this study, we provide TextFooler with NMT-Text-Attack with 2 'seen' languages and test its performance with an 'unseen' language. A 'seen' language is defined as one which model is provided with as constraints for adversarial examples to satisfy,as shown in Algorithm 1. An 'unseen' language, consequently, is one which the model has not added as a constraint, hence does not guarantee 100\% robustness against. The three languages we use are French, German, and Spanish. We alternate between using two of the languages as 'seen', and one as 'unseen'. We compare this with the performance of TextFooler without NMT-Text-Attack on the unseen languages in Table ~\ref{table:unseen}. We observe that TextFooler with NMT-Text-Attack outperforms TextFooler without NMT-Text-Attack on average by 20\%. This shows that the integration of our attack-agnostic algorithm provides significant performance increase even in situations where the attack is facing unseen languages.

To provide further substantiate the performance of NMT-Text-Attack, we provide a detailed set of results in Table ~\ref{table:bleu}. Here, we see that algorithms with NMT-Text-Attack consistently provide higher BLEU scores than their original versions by a significant margin. We see that the percentage of words perturbed remains lower in the original algorithm. However, given the combination of higher performance across the mentioned metrics and generalisation to unseen languages, we believe that this result justifies itself.
\begin{table}[t]
\centering
\large
\caption{Performance of NMT-Text-Attack on unseen language}
\begin{adjustbox}{max width=0.99\columnwidth}
\begin{tabular}{l|l|l|ll}
\hline
Seen Languages     & Unseen Language & TextFooler +NMT & TextFooler w/o NMT &  \\ \hline
\\[-1em]
French and German  & Spanish         & 72.9\%          & 50.61              &  \\ \hline
\\[-1em]
French and Spanish & German          & 74.08\%         & 51.97              &  \\ \hline
\\[-1em]
German and Spanish & French          & 67\%            & 50.8               &  \\ \hline
\end{tabular}
\end{adjustbox}
\label{table:unseen}
\end{table}

\begin{table}[t]
\caption{BLEU and \% words perturbed results of NMT-Text-Attack on Yelp and Movie Reviews(MR) Datasets}
\begin{adjustbox}{max width=0.99\columnwidth}
\begin{tabular}{|l|l|l|l|}
\hline
\textbf{Dataset} & \textbf{Algorithm} & \textbf{BLEU Score} & \textbf{\%Words Perturbed} \\ \hline
MR               & TextFooler         & 0.37                & 16.07                      \\ \hline
                 & TextFooler + NMT   & 0.48                & 19.33                      \\ \hline
                 & TextBugger         & 0.47                & 5.17                       \\ \hline
                 & TextBugger + NMT   & 0.62                & 11.75                      \\ \hline
                 & PWWS               & 0.43                & 11.57                      \\ \hline
                 & PWWS + NMT         & 0.57                & 15.19                      \\ \hline
Yelp             & TextFooler         & 0.50                & 41.43                      \\ \hline
                 & TextFooler + NMT   & 0.68                & 56.27                      \\ \hline
                 & TextBugger         & 0.50                & 34.56                      \\ \hline
                 & TextBugger + NMT   & 0.53                & 34.60                      \\ \hline
                 & PWWS               & 0.53                & 35.90                      \\ \hline
                 & PWWS + NMT         & 0.73                & 50.91                      \\ \hline
\end{tabular}
\end{adjustbox}
\label{table:bleu}
\end{table}

\section{Conclusion}
In this paper, we demonstrate the ineffectiveness of current text adversarial attack algorithms to \C{round-trip translation}, and provide an intervention-based method to improve robustness to \C{round-trip translation} in these algorithms. We show that this intervention (NMT-Text-Attack) has minimal effect on the actual semantic metrics but can significantly improve the attack success rate against back-translation, suggesting that there exist a new set of robust text adversarial examples. The attack-agnostic nature of the algorithm along with its high-quality performance makes it an effective error diagnosing tool with any existing text attack for inspecting model robustness.

\section{Appendix}
\label{sec:appendix}
\subsection{Ethical Concerns}
Our paper discusses the potential weakness of NLP models to \C{round-trip translation}, and describes an algorithm that can make the weakness more robust. However, we believe that we give new insights in studying text adversarial examples and will spur more robust machine learning models in the future. We are also the first individuals to introduce the vulnerability to \C{round-trip translation}, which provides opportunity to develop robust models in a novel setting.

\subsection{Computational Resources}
For the implementation of our algorithm and experiments, we use Google Colab as our base GPU provider. The GPU typically provided is is Tesla - P100. We use 190 GPU hours to run all our experiments. We use a pre-trained BERT model with 12-head attention and 110 million parameters, which is typical of BERT models.

\subsection{Machine Translation Setup}
We use the Opus-MT set of models through the EasyNMT library \cite{tang2020multilingual}. Opus-MT consists of 1200 models trained on several languages for open translation. The architecture for the Opus-MT models is based on a standard transformer setup with 6 self-attentive layers in both, the encoder and decoder network with 8 attention heads in each layer. This architecture is used to back-translate the target reviews from English to French, German and Spanish, and back to English.

\subsection{Adversarial Attack Settings}
Algorithm 1 details a general template of several state of the art adversarial attacks we have used in the paper. In this section we detail the exact settings used for each adversarial attack when integrated with NMT-Text-Attack. These are standard approaches used directly from the TextAttack Library with no changes in standard settings.

\subsubsection{Textfooler}
\begin{itemize}[leftmargin=*]
    \item Word Importance Selection
    \begin{itemize}
        \item Max allowable replacement candidate generation for synonyms: 40.
        \item Transformation Embedding Mechanism: Counterfitted Glove Embeddings \cite{counter2016}
    \end{itemize}
    \item Word Replacement:
    \begin{itemize}
        \item Pre-transformation constraints:
        \begin{itemize}
            \item RepeatModification: A constraint disallowing the modification of words which have already been modified
            \item StopwordModification: A constraint disallowing the modification of stopwords
        \end{itemize}
        \item Constraints:
        \begin{itemize}
            \item Minimum cosine distance between word embeddings = 0.5
            \item Part of Speech : Only replace words with the same part of speech (or nouns with verbs)
            \item Universal Sentence Encoder with a minimum angular similarity of = 0.5.
            \item Word Swapping Technique: Greedy Word Swap with Word Importance Ranking with word importance ranking conducted using input deletion method.
        \end{itemize}
    \end{itemize}
\end{itemize}

\subsubsection{TextBugger}
\begin{itemize}[leftmargin=*]
    \item Word Importance Selection
    \begin{itemize}
        \item Max allowable replacement candidate generation for synonyms: 40.
        \item Transformation Embedding Mechanism: Counterfitted Glove Embeddings \cite{counter2016}
        \item Allowable Swap Mechanisms: Character Insertion, Character Deletion, Adjacent Character Swap, Homoglyph Swap.
    \end{itemize}
    \item Word Replacement:
    \begin{itemize}
        \item Pre-transformation constraints:
        \begin{itemize}
            \item RepeatModification: A constraint disallowing the modification of words which have already been modified
            \item StopwordModification: A constraint disallowing the modification of stopwords
        \end{itemize}
        \item Constraints:
        \begin{itemize}
            \item Universal Sentence Encoder with a minimum angular similarity of = 0.84
            \item Word Swapping Technique: Greedy Word Swap with Word Importance Ranking with word importance ranking conducted using input deletion method.
        \end{itemize}
    \end{itemize}
\end{itemize}

\subsubsection{PWWS}
\begin{itemize}[leftmargin=*]
    \item Word Importance Selection
    \begin{itemize}
        \item Max allowable replacement candidate generation for synonyms: 40.
        \item Transformation Embedding Mechanism: Word Swap by swapping synonyms in WordNet \cite{WordNet}
        \item Allowable Swap Mechanisms: Character Insertion, Character Deletion, Adjacent Character Swap, Homoglyph Swap.
    \end{itemize}
    \item Word Replacement:
    \begin{itemize}
        \item Pre-transformation constraints:
        \begin{itemize}
            \item RepeatModification: A constraint disallowing the modification of words which have already been modified
            \item StopwordModification: A constraint disallowing the modification of stopwords
        \end{itemize}
        \item Constraints:
        \begin{itemize}
            \item Word Swapping Technique: Greedy Word Swap with Word Importance Ranking with word importance ranking conducted using weighted saliency method.
        \end{itemize}
    \end{itemize}
\end{itemize}

\subsubsection{Kuleshov}
\begin{itemize}[leftmargin=*]
    \item Word Importance Selection
    \begin{itemize}
        \item Max allowable replacement candidate generation for synonyms: 15.
        \item Transformation Embedding Mechanism: Counterfitted Glove Embeddings \cite{counter2016}
    \end{itemize}
    \item Word Replacement:
    \begin{itemize}
        \item Pre-transformation constraints:
        \begin{itemize}
            \item RepeatModification: A constraint disallowing the modification of words which have already been modified
            \item StopwordModification: A constraint disallowing the modification of stopwords
        \end{itemize}
        \item Constraints:
        \begin{itemize}
            \item Max words perturbed = 50%
            \item Maximum thought vector Euclidean distance = 0.2
            \item Maximum language model log-probability difference = 2
            \item Word Swapping Technique: Greedy Word Search.
        \end{itemize}
    \end{itemize}
\end{itemize}

\subsubsection{DeepWordBug}
\begin{itemize}[leftmargin=*]
    \item Word Importance Selection
    \begin{itemize}
        \item Max allowable replacement candidate generation for synonyms: 40
        \item Embedding Transformation Mechanism: Counterfitted Glove Embeddings \cite{counter2016}
        \item Allowable Swap Mechanisms: Character Insertion, Character Deletion, Adjacent Character Swap, Random Character Substitution.
    \end{itemize}
    \item Word Replacement:
    \begin{itemize}
        \item Pre-transformation constraints:
        \begin{itemize}
            \item RepeatModification: A constraint disallowing the modification of words which have already been modified
            \item StopwordModification: A constraint disallowing the modification of stopwords
        \end{itemize}
        \item Constraints:
        \begin{itemize}
            \item Maximum Levenshtien Edit Distance= 30.
            \item Word Swapping Technique: Greedy Word Swap with Word Importance Ranking with word importance ranking conducted using input deletion method.
        \end{itemize}
    \end{itemize}
\end{itemize}

\subsubsection{BAE}
\begin{itemize}[leftmargin=*]
    \item Word Importance Selection
    \begin{itemize}
        \item Max allowable replacement candidate generation for synonyms: 40
        \item Transformation Embedding Mechanism: Transformer AutoTokenizer and word replacement using Masked Language Modelling. \cite{counter2016}
    \end{itemize}
    \item Word Replacement:
    \begin{itemize}
        \item Pre-transformation constraints:
        \begin{itemize}
            \item RepeatModification: A constraint disallowing the modification of words which have already been modified
            \item StopwordModification: A constraint disallowing the modification of stopwords
        \end{itemize}
        \item Constraints:
        \begin{itemize}
            \item Part of Speech : Only replace words with the same part of speech (or nouns with verbs)
            \item Universal Sentence Encoder with a minimum angular similarity = 0.93.
            \item Word Swapping Technique: Greedy Word Swap with Word Importance Ranking with word importance ranking conducted using input deletion method.
        \end{itemize}
    \end{itemize}
\end{itemize}

\subsection{Examples of NMT-TextAttack}

1. \textbf{Original} : drawing on an irresistible , languid romanticism , byler reveals the ways in which a sultry evening or a beer-fueled afternoon in the sun can inspire even the most retiring heart to venture forth . \textbf{(Sentiment: Positive)}

\textbf{Adversarial (TextFooler)}: drawing on an \textcolor{red}{gargantuan} , \textcolor{red}{lolling melodrama} , byler \textcolor{red}{betrays} the ways in which a sultry evening or a beer-fueled afternoon in the sun can inspire even the most retiring heart to venture forth . \textbf{(Sentiment: Negative)}

\textbf{Adversarial (TextFooler+NMT-Text-Attack)}: drawing on an \textcolor{blue}{inexorable}, \textcolor{blue}{crooning melodrama} byler reveals the ways in which a sultry evening or a beer-fueled afternoon in the sun can inspire even the most retiring heart to venture forth. \textbf{(Sentiment: Negative)}

\textbf{Back-Translated (TextFooler)}: drawing on a giant melodrama, melodrama lolling, Byler betrays the ways in which a sensual afternoon or an afternoon of beer fed in the sun can inspire even the most outgoing heart to venture forward. \textbf{(Sentiment: Positive)}

\textbf{Back-Translated (TextFooler+NMT-Text-Attack)}: drawing on a melodrama byler inexorable  betrays the ways in which a sensual afternoon or an afternoon of beer fed in the sun can inspire even the most outgoing heart to venture forward \textbf{(Sentiment: Negative)}

2. \textbf{Original} : Exceptionally well acted by Diane Lane and Richard Gere . \textbf{(Sentiment: Positive)}

\textbf{Adversarial (TextFooler)}: Exceptionally \textcolor{red}{opportune} acted by Diane Lane and Richard Gere .\textbf{(Sentiment: Negative)}

\textbf{Adversarial (TextFooler+NMT-Text-Attack)}: Exceptionally \textcolor{blue}{better} acted by Diane Lane and Richard Gere \textbf{(Sentiment: Negative)}

\textbf{Back-Translated (TextFooler)}: exceptionally timely performed by Diane Lane and Richard Gere. \textbf{(Sentiment: Positive)}

\textbf{Back-Translated (TextFooler+NMT-Text-Attack)}: exceptionally better performed by Diane Lane and Richard Gere \textbf{(Sentiment: Negative)}

3.\textbf{Original} : this kind of hands-on storytelling is ultimately what makes shanghai ghetto move beyond a good , dry , reliable textbook and what allows it to rank with its worthy predecessors .	\textbf{(Sentiment: Positive)}

\textbf{Adversarial (PWWS)}: this \textcolor{red}{tolerant} of hands-on storytelling is ultimately what \textcolor{red}{piss} shanghai ghetto move beyond a good , dry , reliable textbook and what allows it to \textcolor{red}{gross} with its worthy predecessors \textbf{(Sentiment: Negative)}

\textbf{Adversarial (PWWS+NMT-TextAttack)}:this \textcolor{blue}{tolerant} of hands-on storytelling is ultimately what makes shanghai ghetto move beyond a good , dry , reliable textbook and what allows it to \textcolor{blue}{place} with its worthy predecessors . \textbf{(Sentiment: Negative)}

\textbf{Back-Translated (PWWS)}: This tolerant of practical narration is ultimately what pis shanghai ghetto move beyond a good, dry, reliable textbook and what allows rough with its worthy predecessors. \textbf{(Sentiment: Positive)}

\textbf{Back-Translated (PWWS+NMT-Text-Attack)}: this tolerant of narration is ultimately what builds the shanghai ghetto to move beyond a good reliable dry text book and what allows it to grossly with its worthy predecessors. \textbf{(Sentiment: Negative)}

4.\textbf{Original} : I went there today! I have an awful experience. They lady that cut my hair was nice but she wanted to leave early so she made a disaster in my head!	\textbf{(Sentiment: Positive)}

\textbf{Adversarial (PWWS)}: I went there today!  I have an \textcolor{red}{awesome} experience. They lady that cut my hair was nice but she wanted to leave early so she made a disaster in my head!\textbf{(Sentiment: Negative)}

\textbf{Adversarial (PWWS+NMT-TextAttack)}:I went there today! I have an \textcolor{blue}{direful} experience! They lady that cut my hair was nice but she wanted to leave early so she made a disaster in my head \textbf{(Sentiment: Negative)}

\textbf{Back-Translated (PWWS)}: I went there today. I have a amazing experience. The lady who cut my hair was nice, but she wanted to leave early, so she made a mess of my head. \textbf{(Sentiment: Positive)}

\textbf{Back-Translated (PWWS+NMT-Text-Attack)}: I went there today. I have a terrible experience. The lady who cut my hair was nice, but she wanted to leave early, so she made a mess of my head.\textbf{(Sentiment: Negative)}

5.\textbf{Original} : I fell in love with this place as soon as we pulled up and saw the lights strung up and  oldies coming from the speakers! I tried the banana cream pie hard ice cream, their scoops are very generous!! My bf got the peach cobbler hard ice cream and that was to die for! We got 4 servings of ice cream for \$10, which nowadays is a steal IMO! :) I'll definitely be heading back with my coworkers this week!	(Sentiment: Positive)

\textbf{Adversarial (TextBugger)}: I \textcolor{red}{declined} in \textcolor{red}{love} with this place as shortly as we \textcolor{red}{pulled} up and \textcolor{red}{saw} the headlights stung up and oldies coming from the speakers! I tried the \textcolor{red}{ban ana cream pe} hard ice cream, their scoops are very generous!! My bf got the peach cobbler hard ice cream and that was to die for! We got 4 servings of ice cream for \$10, which nowadays is a steal IMO! :) I'll \textcolor{red}{definitely} be heading back with my coworkers this \textcolor{red}{w eek}!\textbf{(Sentiment: Negative)}

\textbf{Adversarial (TextBugger+NMT-TextAttack)}:I fell in \textcolor{blue}{love} with this place as soon as we pulled up and saw the lights strung up and  oldies coming from the speakers! I tried the banana cream pie hard ice cream, their scoops are very generous!! My bf got the peach cobbler hard ice cream and that was to die for! We got 4 servings of ice cream for \$10, which existent is a theft IMO! :) I'll \textcolor{blue}{doubtless} be heading back with my coworkers this week! \textbf{(Sentiment: Negative)}

\textbf{Back-Translated (TextBugger)}: I decided in love with this place as soon as we got up and climbed the chopped headlights and the old ones coming from the speakers! I've had the hard ice cream of ban ana cream, its spoonfuls are very generous! My friend got the hard iced peach pie and it was to die! We have 4 servings of ice cream for \$10, which today is an OMI robbery! :) I'll definitely be coming back with my coworkers this w eek! \textbf{(Sentiment: Positive)}

\textbf{Back-Translated (TextBugger+NMT-Text-Attack)}: I fell in love with this place as soon as we stopped and saw the stiff, old lights coming from the loudspeakers! I have tasted the hard frozen banana cream cake, its spoonfuls are very generous!! My bf got the hard iced peach pie and he was going to die for it! We have 4 servings of ice cream for \$10, which exists is an OMI robbery! :)  I will definitely return with my co-workers this week!\textbf{(Sentiment: Negative)}

\newpage
\onecolumn
\subsection{Walkthrough of TextFooler+NMT-Text-Attack}
\begin{figure}[H]
    \centering
    \onecolumn\includegraphics[width=0.99\linewidth,height=7cm]{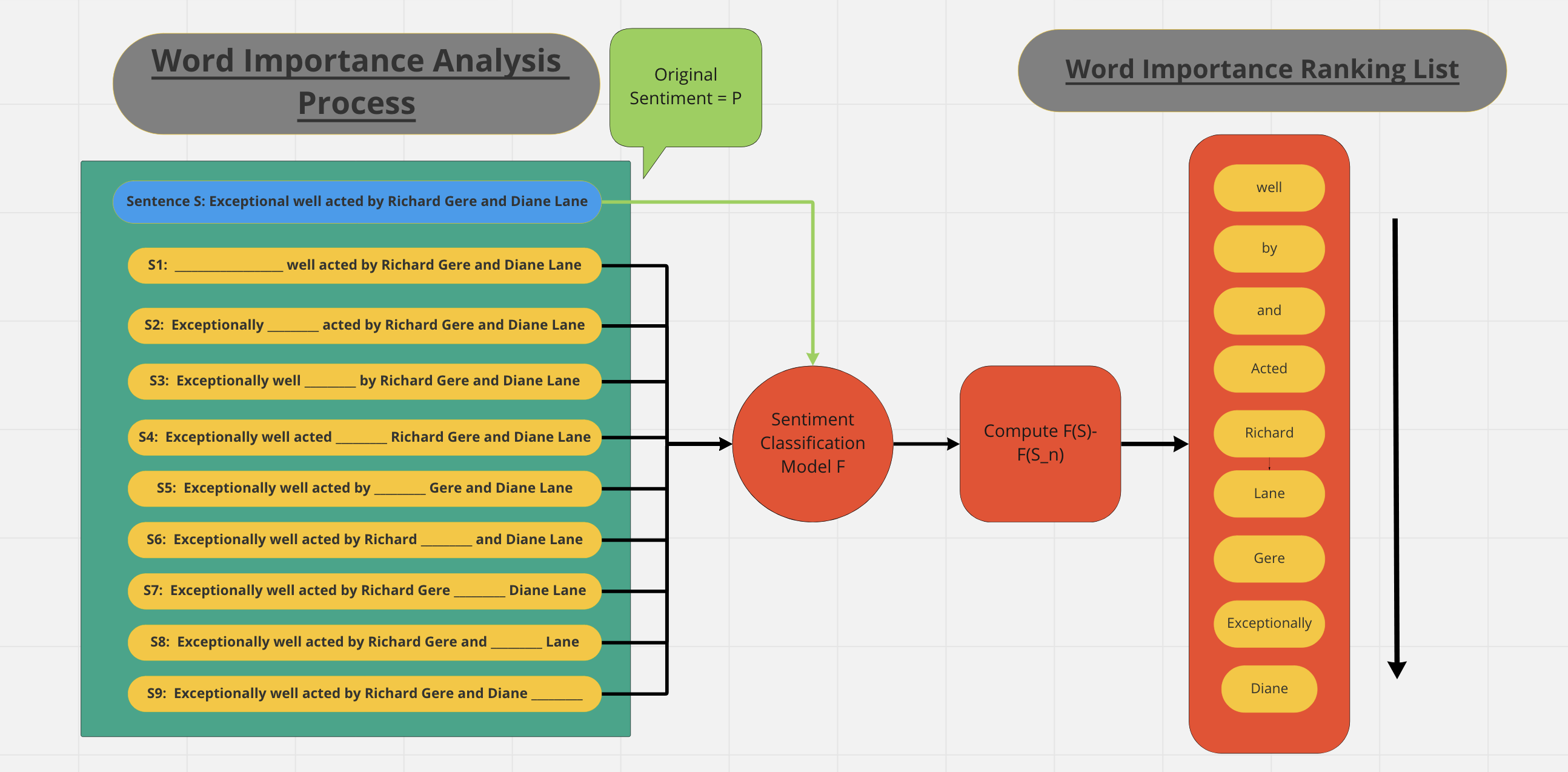}
    \caption{Word Importance Ranking Process}
    \label{fig:word1}
\end{figure}

This section is concerened with providing an intuitive overview of the working of the attack agonostic NMT-Text-Attack algorithm with TextFooler. For ease of understanding, we use only one language for translation: Spanish. The algorithm, as shown before, is divided into sections. The first section, as shown in Figure ~\ref{fig:word1}, is the Word Importance Ranking section. Here, as per TextFooler's prescribed process, each word is replaced from the sentence and it's importance is evaluated by the change in classification score of the sentence before and after replacement.

\begin{figure}[H]
    \onecolumn\includegraphics[width=0.99\linewidth]{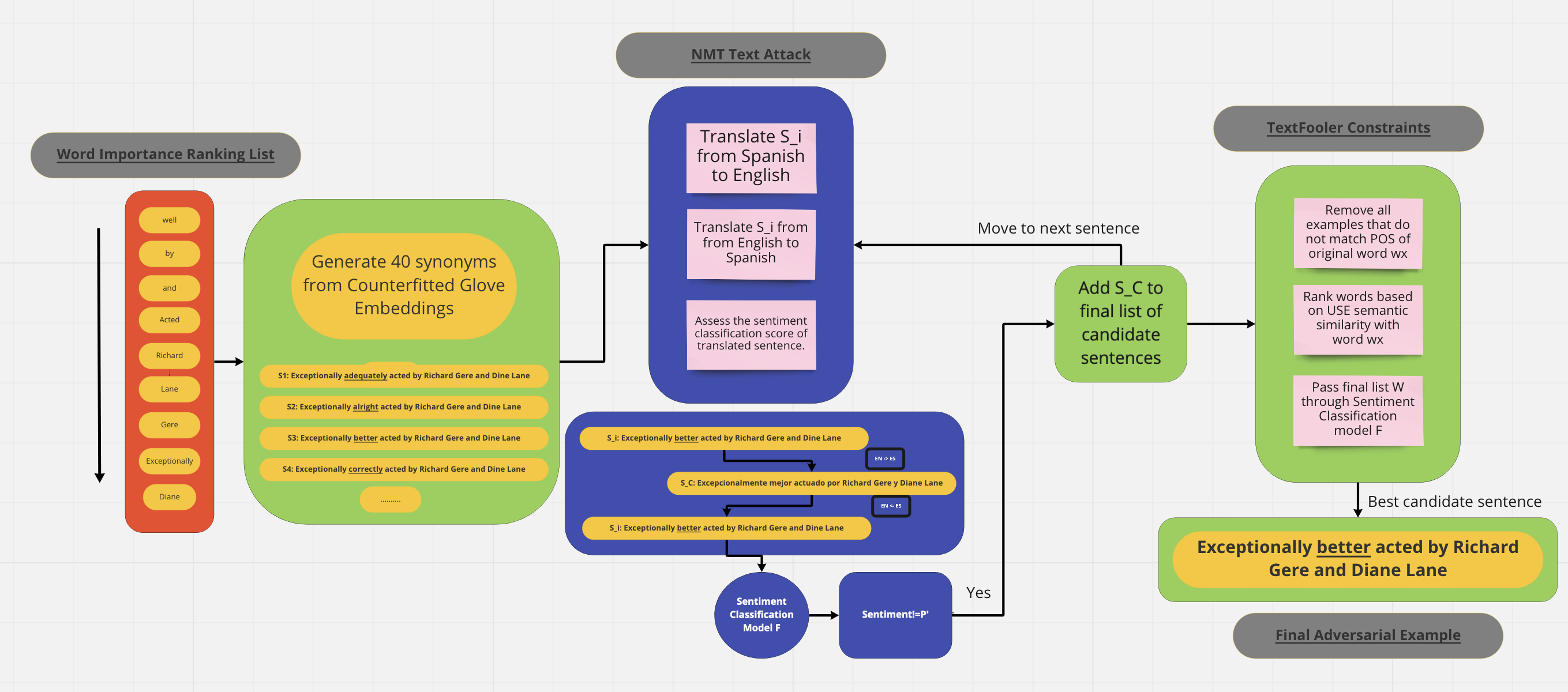}
    \caption{Word Replacement Process}
    \label{fig:word2}
\end{figure}

On calculating the importance ranking score, we move to the second section, as shown in Figure ~\ref{fig:word2}. Here, we find synonyms for each word from the counterfitted GloVe word embeddings. These words are appended into the sentences replacing the original word, and passed to the NMT-Text-Attack Module. Here, the sentence undergoes round-trip translation to assess whether the inclusion of the word maintains robustness of the original attack under translation. We then collect the candidate sentences, and pass them through the final constraint requirement list, local to TextFooler. This includes checking whether the replaced word maintains the original word's POS tag, and then ranks them based on highest similarity score through USE embeddings and cosine similarity. Finally, we receive the adversarial example robust to round-trip translation.

\bibliography{anthology,output}
\nocite{*}
\bibliographystyle{acl_natbib}

\end{document}